\documentclass{article}

\PassOptionsToPackage{numbers,compress}{natbib}

\usepackage[preprint]{neurips_2025}

\usepackage[utf8]{inputenc}
\usepackage[T1]{fontenc}

\usepackage{hyperref}
\usepackage{url}
\usepackage{booktabs}
\usepackage{amsfonts,amsmath}
\usepackage{nicefrac}
\usepackage{microtype}
\usepackage{multirow}
\usepackage{makecell}
\usepackage{graphicx}
\usepackage{subcaption}
\usepackage{todonotes}
\usepackage{placeins}
\usepackage{CJKutf8}
\usepackage{booktabs}
\usepackage{colortbl}
\usepackage{framed}
\usepackage{listings}
\usepackage{xcolor}
\usepackage{float}

\definecolor{lavender}{RGB}{204,204,255}

\title{%
  MedBrowseComp: Benchmarking Medical Deep Research and Computer Use%
}

\author{%
  Shan Chen$^{1,2,3}$\thanks{Co-first authors: Shan Chen and Pedro Moreira},
  Pedro Moreira$^{4,5}$\footnotemark[1],
  Yuxin Xiao$^{4}$\\
  \textbf{Sam Schmidgall$^{6}$, Jeremy Warner$^{7,8}$, Hugo Aerts$^{1,2,9}$}\\
  \textbf{Thomas Hartvigsen$^{10}$, Jack Gallifant$^{1,2}$,
  Danielle S. Bitterman$^{1,2,3,8}$\thanks{Corresponding author: dbitterman@bwh.harvard.edu}}\\
  \\
  $^1$Harvard, $^2$Mass General Brigham, $^3$Boston Children's Hospital, $^4$MIT,\\
  $^5$Universitat Pompeu Fabra, $^6$Johns Hopkins University, $^7$Brown University,\\
  $^8$HemOnc.org, $^9$Maastricht University, $^{10}$University of Virginia\\
}
\def\huggingface{\raisebox{-1.5pt}{\includegraphics[height=1.05em]{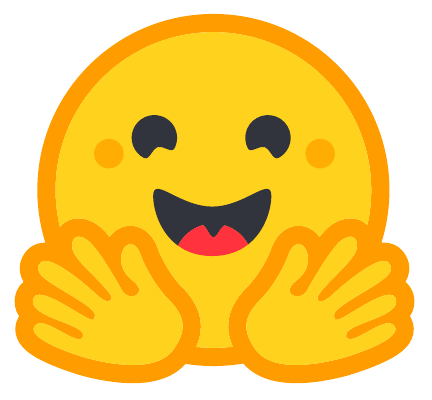}}}
\def\github{\raisebox{-1.5pt}{\includegraphics[height=1.05em]{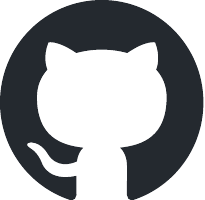}}}
\newcommand{\hflink}{https://huggingface.co/datasets/AIM-Harvard/MedBrowseComp}
\newcommand{\ghlink}{https://github.com/shan23chen/MedBrowseComp}
\begin{document}
\maketitle
\begin{center}
\vspace{-1cm}
\begin{tabular}{rl}
\huggingface & \url{\hflink}\\
\github & \url{\ghlink}\\
\end{tabular}
\end{center}
% --------------------------- Abstract -------------------------------------
\begin{abstract}
Large language models (LLMs) are increasingly envisioned as decision-support tools in clinical practice, yet safe clinical reasoning demands the integration of heterogeneous knowledge bases—trials, primary studies, regulatory documents, and cost data— under strict accuracy constraints. Existing evaluations typically rely on synthetic prompts, reduce the task to single-hop factoid queries, or conflate reasoning with open-ended text generation, leaving their real-world utility unclear. To close this gap, we present \textbf{MedBrowseComp}, the first benchmark that systematically tests an agent’s ability to reliably retrieve and synthesize multi-hop medical facts from live, domain-specific knowledge bases. MedBrowseComp holds 1,000+ human-curated questions that mirror clinical scenarios in which practitioners must reconcile fragmented or conflicting information to reach an up-to-date conclusion. Applying MedBrowseComp to frontier agentic systems reveals \textbf{marked performance shortfalls as low as 10\%}. These findings expose a critical gap between current LLM capabilities and the rigor demanded in clinical settings. MedBrowseComp exposes the strengths and weaknesses of current agentic systems, offering a testbed for reliable medical information seeking and clear goals for future model and toolchain upgrades.
\end{abstract}

% --------------------------- Main sections --------------------------------
\section{Introduction}% Motivation
LLMs have now effectively saturated most static, knowledge-based benchmarks, diminishing the value of these tasks to provide new insights and push the field forward\cite{guha2023legalbench,hendrycksmath2021,hendrycks2021measuring,liu2023benchmarking,ye2024qilinmed,phan2025humanitysexam}. However, this evolution exposes an \emph{evaluation gap}: legacy leaderboards track static knowledge recall, whereas agentic systems should \textit{plan}, \textit{browse}, and \textit{synthesize} fresh evidence in real time. To move beyond this plateau, the community is pivoting toward \emph{agentic} systems that actively browse the web, retrieve real-time evidence, and reason over information outside their frozen parameter memories \cite{lewis2020retrieval,vu2023freshllms,alzubi2025open}. The progression from chatbots to reasoners and ultimately to autonomous agents promises to enable access to real-time data, to allow models to tackle questions outside their pretraining knowledge, and perform complex information gathering tasks previously exclusive to humans \cite{nakano2022webgptbrowserassistedquestionansweringhuman,wang2023voyageropenendedembodiedagent,kasai2024realtimeqawhatsanswer}.

% State-of-the-art and knowledge gap
The potential impact of web-enabled agents is enormous. In principle, a sufficiently-capable AI agent should be able to retrieve any well-specified fact from the open web, even if doing so requires navigating thousands of pages \cite{deng2023mind2web,xu2024aguvis,pahuja2025explorer}. This promise is especially compelling in medicine, where research, clinical decision support, and patient education all depend on the integration of the most current, specific, and accurate information from heterogeneous sources such as journal articles, clinical trial registries, treatment guidelines, and drug databases \cite{wornow2023shaky,chen2023impact,chen2023use,zhou2024survey,yu2024medical,tu2024conversationaldiagnosticai,gallifant2025tripod}. Yet despite rapid progress, the community lacks a unified benchmark for systematically evaluating whether agents can perform such complex, multi-source medical retrieval at scale.

As LLMs become the backbone of autonomous agentic systems, rigorous, domain-specific evaluation is critical. Contemporary agent models frequently “hallucinate,” generating confident but unsubstantiated or factually incorrect statements \cite{kim2025medical}. In high-stakes fields like medicine, these errors can misinform clinicians or patients and erode trust. Therefore, benchmarks must test not only a model’s reasoning and navigation skills but also its ability to ground each answer in verifiable evidence \cite{chen2023use,zhou2024survey,kim2025medical}. A robust framework that measures evidence-based accuracy, navigation efficiency, and citation fidelity would directly quantify how well agentic systems incorporate best available information, closing the gap between impressive demos and safe, real-world deployment.

Evaluating an agent’s deep research and computer use competence is fundamentally different from testing its ability to recall static facts. Popular medical benchmarks such as MMLU, MedQA, and WorldMedQA test information that can be memorized during pre-training or retrieved from a single authoritative page. Frontier models now achieve near-ceiling scores on these tasks, so the benchmarks can no longer distinguish state-of-the-art systems or quantify new progress \cite{hendrycks2021measuring,jin2020medqa,matos2024worldmedqavmultilingualmultimodalmedical}. They are usually executed in closed or synthetic environments that sidestep the real-world hurdles of live web interaction—pagination, obsolete links, contradictory evidence, and shifting page layouts \cite{zhang2024android,xie2024osworld}.

 In medicine, clinicians and researchers must integrate the latest study results, guideline updates, and safety advisories scattered across heterogeneous sites. To understand and compare agents'  abilities to augment such tasks, new deep research and computer use benchmarks reflective of real-world, up-to-date tasks are urgently needed. A benchmark that forces agents to conduct multi-hop, evidence-grounded searches on the open Web should therefore serve two complementary purposes: \textbf{1)} Directly measure whether agents can navigate, filter, and reconcile real-world information. \textbf{2)}
Dynamically stress-test systems as underlying evidence evolved, long after static benchmarks saturate. 

To address this gap, we introduce MedBrowseComp, a new benchmark specifically designed to evaluate the capabilities of AI agents performing complex information retrieval tasks within the medical domain via web browsing. MedBrowseComp measures an agent's ability to accurately navigate the web—including general web pages, specialized medical websites, databases, and potentially document formats like wikis, PDFs—to locate verifiable medical facts.

MedBrowseComp's design is inspired by the BrowseComp benchmark \cite{wei2025browsecomp}; it focuses on fact-seeking questions where the answers are short, objective, and easily verifiable, simplifying the evaluation process and enhancing its reliability. \textbf{We designed this challenging benchmark collaboratively with physicians using HemOnc.org, one of the largest structured wiki information resources maintained weekly by oncologists for the past 6 years.} The benchmark is designed to enable automated, dynamic updating as information resources evolve. State-of-the-art Deep Research systems and Computer Use Agents(CUA) in May 2025 achieve less than 50\% accuracy overall on MedBrowseComp, with less than 10\% in the two hardest sets of questions.

The primary contributions of this work are:

\textbf{The MedBrowseComp Dataset:} A novel, curated collection of challenging medical fact-seeking questions, each requiring web browsing and resulting in a short, verifiable answer. The pioneer in curating a comprehensive benchmark that utilizes linked domain knowledge.

\textbf{Baseline Performance Analysis:} An empirical evaluation of various state-of-the-art LLMs and agentic systems on MedBrowseComp, providing initial benchmarks and highlighting the specific difficulties encountered in medical information retrieval.

\textbf{Demonstration of Capability Gaps:} Evidence showing the gap between the capabilities of general-purpose browsing agents and specialized skills on complex medical information-seeking tasks.

\section{Related Work}\begin{figure}[t]
  \centering
  \includegraphics[scale=0.2]{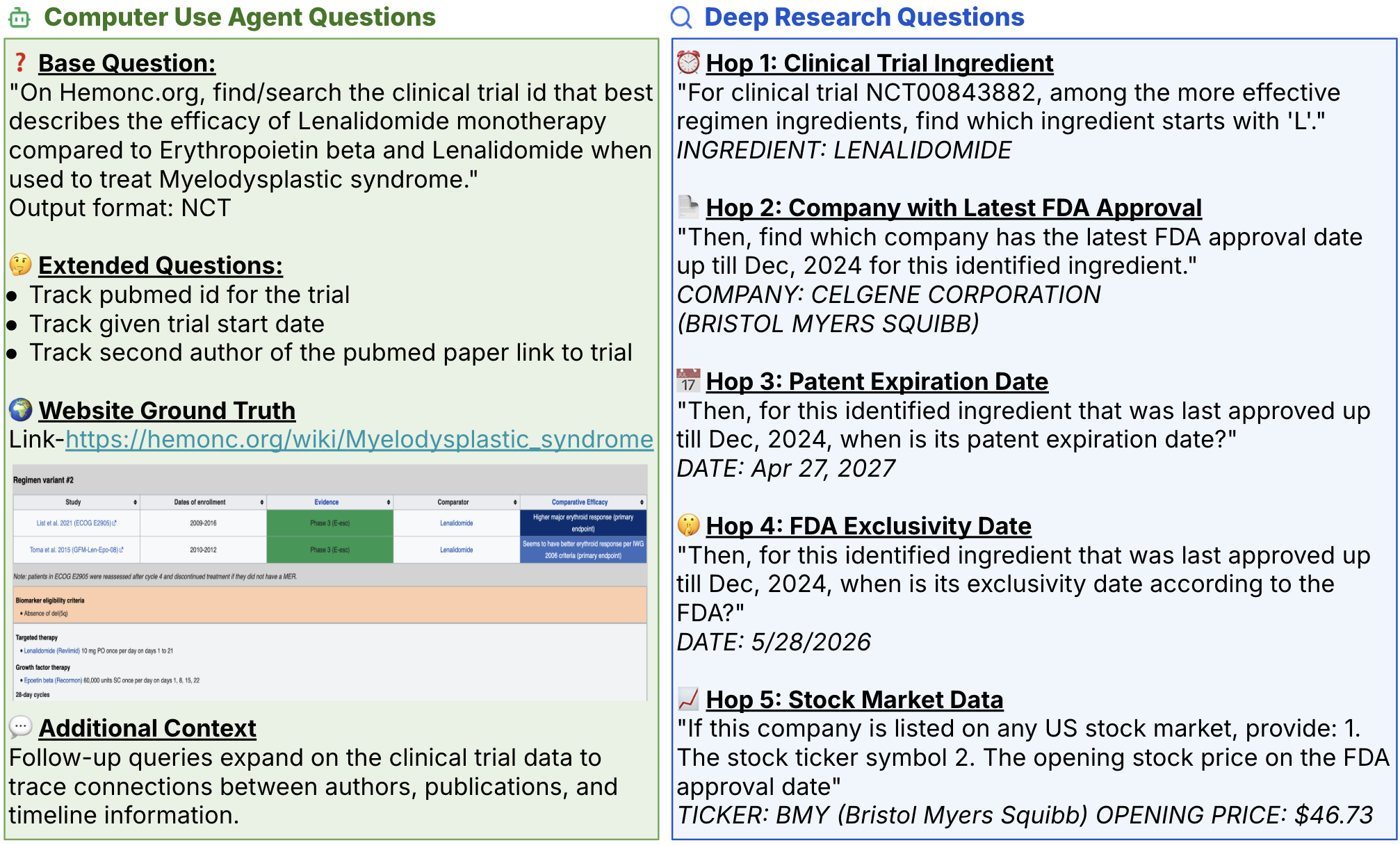}
  \caption{Example question constructions for MedBrowseComp.}
  \label{fig:workflow}
\end{figure}

One of the first comprehensive efforts to advance AI evaluation for general-purpose tool use and web browsing is GAIA \cite{mialon2023gaia}. GAIA is a carefully curated testbed for “General AI Assistants,” combining tasks that require multi-modal input, open-ended reasoning, and the ability to query external tools. It is simple for humans, but it was hard for AI systems at the time. Subsequent work began to highlight the importance of structured, multi-step web traversal. WebWalker introduced a dual-agent framework that separately handles horizontal browsing across different webpages and deep vertical navigation through website hierarchies \cite{wu2025webwalker, xie2024osworld}. Its companion benchmark, WebWalkerQA, comprises realistic multi-hop questions from domains like education and organizational websites, emphasizing the challenge of distributing information across intricate hyperlink structures. Around the same time, FRAMES took a different angle by systematically evaluating Retrieval-Augmented Generation (RAG) systems for factual correctness, retrieval quality, and reasoning \cite{krishna2024fact}.  While WebWalker and FRAMES address structured exploration and pipeline analysis, SimpleQA focuses more narrowly on short-answer factual correctness, specifically targeting LLM hallucinations.\cite{wei2024measuring} SimpleQA’s adversarial question collection and stringent answer verification make it an effective tool for measuring whether models can reliably produce grounded, non-invented responses. Despite its value in exposing factual flaws, SimpleQA’s tasks remain relatively easy to solve with basic web searches, limiting its utility for deeper or more specialized queries \cite{alzubi2025open}.

To test expert-level knowledge well beyond the range of average tasks, Humanity's Last Exam (HLE) provides a set of deeply specialized questions \cite{phan2025humanitysexam}. HLE intentionally targets areas where frontier models are known to have knowledge gaps, thereby illuminating the upper limits of AI comprehension across diverse academic disciplines. However, while HLE excels at assessing conceptual difficulty and depth of reasoning, it focuses less on web and knowledge-source navigation capabilities.

BrowseComp and its offshoot BrowseComp-ZH push AI agents to perform complex web searches for difficult-to-find facts \cite{wei2025browsecomp, zhou2025browsecomp}. BrowseComp's 1,255 English questions use a reverse-engineered approach to ensure they are not trivially searchable: each was devised and combined from existing short answers. The leading OpenAI models have an overall accuracy lower than 10\%, however AI systems with search functionalities perform better. BrowseComp-ZH extends this framework to the Chinese web, accounting for different linguistic structures, censorship constraints, and local data sources. Leading models still achieve below 10–20\% on BrowseComp-ZH tasks, suggesting that domain- or region-specific peculiarities are another consideration of the difficulty of high-stakes information retrieval.

Taken together, these benchmarks outline a rapidly evolving space in AI applications, where persistent web navigation, accurate retrieval, and robust reasoning remain underdeveloped in even the most capable models \cite{song2025bearcubs}. MedBrowseComp bridges the gap by focusing on the specialized domain of medicine. Its 1000+ questions\footnote{121*5 -> 605 deep research questions across five hops, 121*4 -> 484 computer use questions over four hops} are designed to be both practically useful and challenging, demanding persistent exploration of reputable medical sources, correct interpretation of terminology, and alignment with evidence-based standards. This unique focus is not only inherently high-stakes but also exposes the limitations of generic benchmarks, which rarely require the same level of detailed domain reasoning or cross-referencing. Moreover, MedBrowseComp’s design is expandable, allowing for continual updates that track evolving medical knowledge and new online resources.

\section{Methods}\begin{figure}[t]
  \centering
  \includegraphics[scale=0.161]{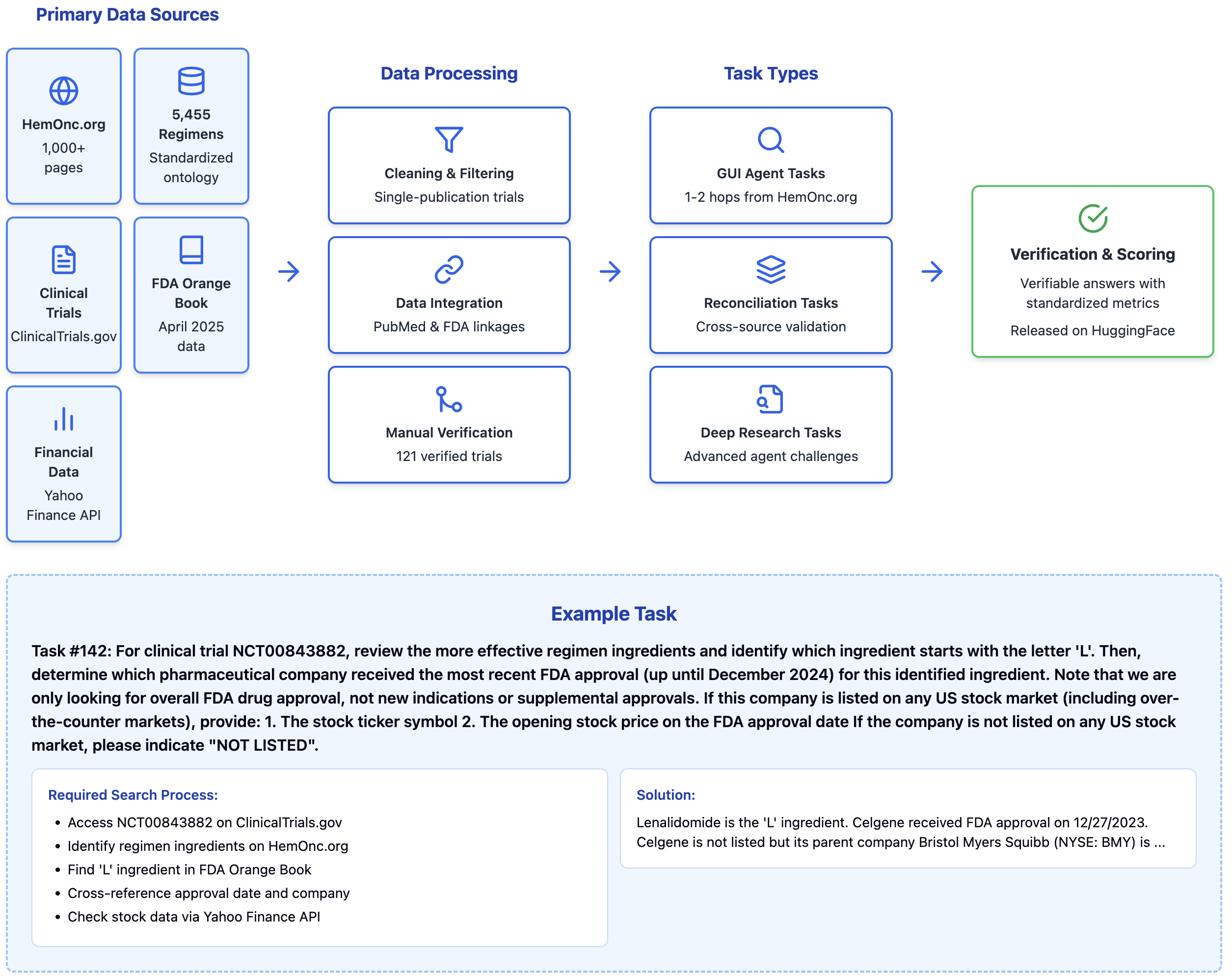}
  \caption{Overall workflow of the curation of MedBrowseComp.}
  \label{fig:workflow}
\end{figure}
To construct the MedBrowseComp dataset, we leverage the comprehensive hematology and oncology database available from HemOnc.org and work with its editors. HemOnc.org is the largest freely available medical wiki in the field of hematology/oncology, established to address the challenge oncologists routinely face navigating complex treatment regimens and rapidly evolving standards of care. This comprehensive resource covers over 1,000 pages of specialized content, including more than 250 hematologic and oncologic conditions, 5,455 detailed treatment regimens, and 6,950 referenced clinical studies, all curated by physicians with verifications. The platform catalogs approved systemic antineoplastic therapy agents, supportive medications, standard-of-care regimens, and references to primary literature, organized within a standardized ontology framework available through the HemOnc Dataverse \cite{Warner2019HemOnc}. Our first step involved cleaning anti-neoplastic regimen efficacy data, linking each case with corresponding PubMed publications, and associated clinical trial information sourced from ClinicalTrials.gov, with data collected up to April 2025. The fully cleaned and structured version of this dataset has been publicly released on HuggingFace to facilitate broader community engagement, further development, and external validation \footnote{https://huggingface.co/datasets/AIM-Harvard/MedBrowseComp\_Meta}. To avoid potential dataset contamination, we encoded our final test sets with shifts and byte-wise encoding, which you can decode using the script we provide on GitHub.

To create our specific evaluation questions, we narrowed our dataset by excluding trials linked to multiple PubMed publications to maintain clarity and verifiability. Subsequently, we integrated regimen-specific drug information with FDA Orange Book data as of April 2025. To maintain data consistency, only trials with regimens containing drugs easily matched through standard generic regular expressions were included. A manual verification and deduplication process, led by author SC, was conducted to ensure accuracy and reduce redundancy, culminating in a refined set of 121 trials. Each of these trials has clearly defined trial metadata, verified regimen efficacy data, detailed FDA drug approval information, and the corresponding financial market data obtained from Yahoo Finance API for associated stock pricing, as Figure \ref{fig:workflow} shows. 

From this carefully curated dataset, we developed our benchmark designed explicitly to assess (1) autonomous CUA within one to two hops of HemOnc.org’s webpage, and deep research agents.

\subsection{Model Selection and other Details}
\paragraph{System Selection Rationale for Deep Research Agent}
We evaluate a range of systems, from models with easy API access and systems without API access, and one Computer Use Agent system. For models with an easy API with accessible cost, we evaluated the full set, which we refer to as MedBrowseComp-605. For models without easy API acccess and/or inaccessible costs, we evaluate against a smaller set which we refer to as MedBrowseComp-50 \footnote{Author SC and JG put each of the queries into each application, copied out the responses, and graded the final outputs.}. Detailed model/system descriptions are in the appendix \ref{bench_details}. For answer verification we employed an automated judge powered by GPT 4.1 mini-2025-04-14. The judging prompt was adapted from existing refined evaluation templates\cite{phan2025humanitysexam, wei2025browsecomp}. Two annotators (SC and JG), manually answered  MedBrowseComp-50 and achieved 100\% inter-annotator agreements and 98\% agreement with GPT 4.1 mini. 

\paragraph{Model Selection Rationale for Computer Use Agent}
We select Anthropic's Computer Use (using Claude 3.7 Sonnet) for evaluating browsing capabilities because, among the major commercially available GUI agents at the time, it was the only one that simultaneously offer a documented, programmable API and sandboxed environment that could be reengineered to run large batches without manual oversight, enabling us to run our experiments at scale.\footnote{Details on Claude's sandbox is provide in the following paragraph and the code is open sourced on our Github repository.}
Other UI agents—such as Bytedance’s UI Tars, OpenAI's Operator, and Google's Project Mariner-were considered, but at the time of experimentation they either lacked comparable logging/automation features or required additional deployment effort that was beyond our project timeline.

\paragraph{Evaluation Architecture}
CUA is still experimental: every UI action spawns a tool call and screenshot, adding range from 2–10k tokens and extra latency. As recent benchmarks and surveys note, this overhead quickly pushes long tasks toward the model’s context limit and amplifies selector failures, making deep navigation chains unreliable. Accordingly, we restrict our benchmark for UI agents to four tasks - trial ID, second author, PMID, and start date — that can be retrieved in one or two hops from HemOnc.org, providing a challenging yet attainable benchmark for GUI agents.

We implement a distributed evaluation infrastructure by adapting Anthropic's CUA codebase, transforming it from an interactive web application into a scalable, fault-tolerant parallel processing system that can be easily evaluated on personal laptops. 

The evaluation back-end is organized as three cooperating services: 
% \begin{enumerate}
%   \item \textbf{Prompt processor} – a lightweight Python microservice that streams CSV-encoded tasks into Claude 3.7 Sonnet, appends a system prompt, and retries up to five times with exponential back-off on transient errors (HTTP 429, timeouts, 5xx).
%   \item \textbf{Container orchestrator} – a Docker/Compose layer that launches $N$ identical Ubuntu 22.04 containers, mounts a shared volume, and shards the prompt file so each container processes an independent batch without crosstalk. We ran our container in a batch of 8 on author PM's MacBook throughout the experiment.
%   \item \textbf{Sampling loop} – issues the model call for each prompt, caps each turn at 4,096 output tokens, and supplies only the three most recent screenshots so the running context never exceeds 120k tokens (well below Sonnet’s 200k window).
% \end{enumerate}
\textbf{(1) Prompt processor} - a lightweight Python microservice that streams CSV encoded tasks into Claude 3.7 Sonnet, applying up to five exponential backoff retries for transient errors (HTTP 429, timeouts, 5xx).  \textbf{(2) Container orchestrator} - a Docker / Compose layer that launches N identical Ubuntu22.04 containers, mounts a shared volume, and shards the prompt file so that each container runs an independent batch without crosstalk. We ran our container in a batch of 8 on author PM's MacBook throughout the experiment. \textbf{(3) Sampling loop} - A sampling loop that calls each turn is capped at 4096 output tokens while the screenshots that are sent as input are only the three most recent, ensuring that the running context never exceeds 120k tokens - well under Sonnet's 200k window.
Note that there is no limit on the number of actions the model can perform or time constraints. We retained Anthropic’s default Ubuntu sandbox exactly as shipped, so the model runs in the same environment used for training — maintaining the best navigation accuracy, reproducibility, and runtime stability. The only change was a minor adjustment to the system‑prompt that nudges the agent to act autonomously without asking a human for grounding or clarifications during execution which you can refer at Appendix \ref{CUA-prompt}. 

Lastly, we verified its difficulty by running Gemini-2.0-Flash-001 zero-shot across five runs, where it achieved under 7 \% accuracy consistently on all sub-partitions of our benchmark questions.

% \paragraph{Implementation Details}
% The system monitors and manages process creation, implementing automatic cleanup routines to prevent resource leakage during long-running evaluation sessions.

% A critical modification to the default system prompt instructed the model to operate autonomously without requesting user clarification, enabling consistent evaluation across all benchmark tasks. This autonomy-focused directive represents an important step toward evaluating models' ability to independently navigate information spaces under uncertainty—a critical capability for deployment in real-world medical contexts. The containerized approach with shared volume architecture facilitated efficient prompt distribution and result aggregation across evaluation instances, while maintaining isolation between concurrent evaluations to prevent cross-contamination.

% \paragraph{Performance Considerations}
% Complex, multi‑step GUI actions (scrolling author lists, switching tabs, copying fields) often mask the agent’s true reasoning errors. To separate these factors, we created a low‑hop benchmark of single‑screen, verifiable queries: each asks for one canonical attribute—NCT ID, start date, second author—obtainable from a single ClinicalTrials.gov or PubMed page. This keeps the biomedical challenge (choosing the correct trial or paper) while minimizing interface friction.Below we report performance on this streamlined task set, with Table\ref{tab:common-errors} highlighting typical failure modes.

\section{Results}%%%% Findings %%%%
\subsection{Deep Research Agent Results}
\begin{figure}[t]
  \centering
  \includegraphics[scale=0.33]{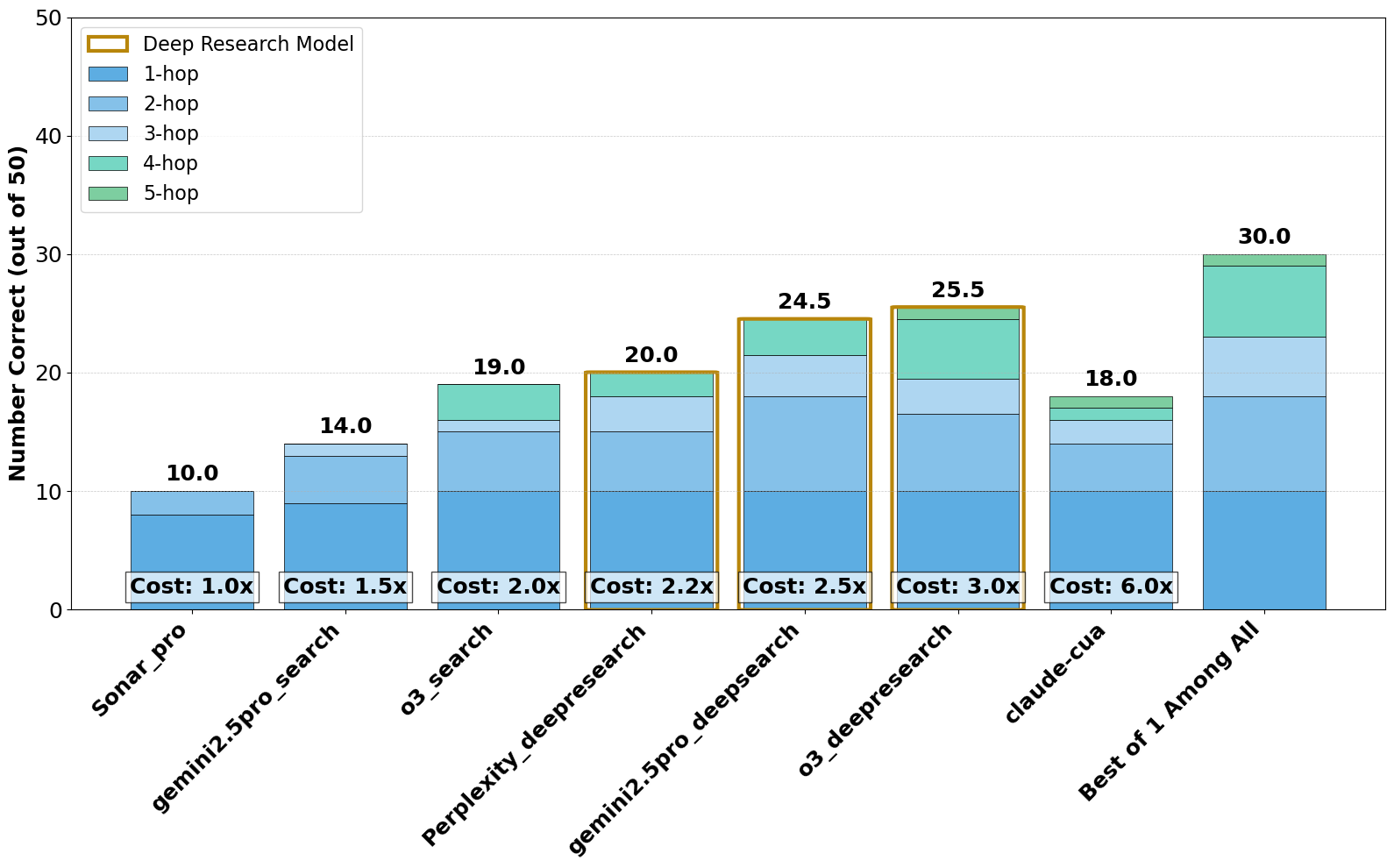}
  \caption{Overall performance of MedBrowseComp. Costs are rough estimations sorted along the x-axis. \textbf{Best of 1 Among All aggregate all measured models and their outputs.} Half point given to a specific 2-hop question where the model retrieved the sub-entities' company name instead.}
  \label{fig:deepresearch}
\end{figure}

Figure \ref{fig:deepresearch} summarizes accuracy on MedBrowseComp-50. Across all systems, performance decays monotonically with hop count, corroborating prior evidence that long-horizon web navigation remains an open challenge for frontier LLM agents. Nevertheless, deep research variants--agents that allow iterative browsing steps rather than a single query—had improved performance. For example, O3 deepresearch answers 25.5/50 questions correctly, a 34\% relative gain over O3 search (19/50); Gemini-2.5-pro deepsearch shows 75\% improvement over its single-shot analogue (24.5 vs. 14). These gains are most pronounced on the hardest 4- and 5-hop splits, where deep research agents more than double the baseline accuracy.

Consistent trends are observed in the MedBrowseComp-605 results, where we exclusively evaluate models utilizing parametric memories and the RAG framework. The performance of bare models is notably poor across the majority of tasks, which aligns with our intention to create a challenging benchmark. RAG improves overall performance, but its benefit diminishes with increasing hops. 
On MedBrowseComp-605, we observe the same core patterns when comparing “bare” parameter-only models—i.e., those relying exclusively on their internal (parametric) memory—with retrieval-augmented variants. In isolation, parameter-only systems struggle across nearly every hop depth, confirming that our benchmark delivers the intended level of difficulty. When doing RAG, models demonstrate substantial gains on shallow questions (1- and 2-hop) except GPT4.1, Gemini2Flash holds 30\% and 4\% boost respectively. And 67\% and 7.4 \% for Gemini2.5Pro. However, the utility of retrieval diminishes beyond the third hop: by the 4- and 5-hop levels, RAG provides virtually no additional benefit over the bare model.
The detailed results of MedBrowseComp-50 and MedBrowseComp-605 are in the Appendix \ref{deepresearch_results}. 

System-wise, O3 deepresearch and Gemini-2.5-pro deepsearch constitute the frontier, trailing only the upper bound of the 'Best of 1' (30/50) that selects post hoc the single best model/system answer per question.\footnote{Unlike prior work showing that repeatedly sampling from a single system can boost performance, our cross-model, test-time compute extension demonstrates even greater gains in overall accuracy \cite{wei2025browsecomp}. However, the computational expense of querying multiple distinct agents for every question is substantial, underscoring the urgent need to develop more efficient, unified systems that deliver comparable or superior results with lower resource overhead.} Their advantage over specialized retrieval systems such as Sonar Pro (10/50) or Perplexity deepresearch (20/50) may suggest that contemporary instruction-tuned LLMs can outperform purpose-built agentic pipelines when granted autonomous browsing. However, even the best system falls short of perfect accuracy, underscoring the need for research in planning, tool use, and hallucination suppression in complex biomedical information-seeking tasks. Table \ref{tab:common-errors} shows some common error modes in examples. %(1) \textbf{Inefficient tool allocation}—the agent spends its quota on low-value checks and exhausts calls before reaching the high-latency finance API that decides the answer; (2) \textbf{Poor source selection}—it relies on press releases instead of the FDA \textit{Orange Book}, letting endpoint errors cascade into incorrect approval or exclusivity dates; and (3) \textbf{Fragile table parsing}—multi-page PDF tables defeat its generic parsers, producing plausible yet incomplete extractions. Together, these missteps call for dynamic tool budgeting, domain-aware source whitelists with freshness checks, and lightweight post-validators—without them, even large-context agents will continue to stumble on multi-hop biomedical queries. 

\begin{table}[t]
\centering
\caption{Examples of common errors among deep research systems on MedBrowseComp‑50.}
\label{tab:common-errors}
\begin{tabular}{p{0.45\textwidth}p{0.5\textwidth}}
\toprule
\textbf{Error type \& question (paraphrased)} & \textbf{Explanation} \\
\midrule
\textbf{Type:} \colorbox{lavender}{Inefficient tool allocation}
\newline\newline
\textbf{Question:} “For clinical trial \texttt{NCT00974311}, review the more effective regimen ingredients and identify which ingredient starts with the letter ‘E’. Then, determine which pharmaceutical company received the most recent FDA approval (until December 2024) for this identified ingredient. If this company is listed on any US stock market, provide the stock ticker symbol and opening stock price on the FDA approval date.” 
& Agent exhausted its tool‑call quota on preliminary tasks (trial verification, news validation) instead of reserving sufficient calls for retrieving critical financial data (stock price lookup). Consequently, the final essential query regarding the stock price was left unanswered, with the agent stating financial data was unavailable in the provided materials.\\
\midrule
\textbf{Type:} \colorbox{lavender}{Poor source selection}
\newline\newline
\textbf{Question:} “In trial \texttt{NCT00974311}, identify the effective ingredient beginning with ‘E’ and return its FDA exclusivity date (overall approval only, in \texttt{MM‑DD‑YYYY} format).”
& The agent cited secondary press releases and FDA news items (e.g., Pfizer announcements) instead of querying the FDA \textit{Orange Book}, the authoritative source for exclusivity information. As a result, it either reported the initial approval date rather than the exclusivity expiry or claimed the exclusivity data were unavailable.\\
\midrule
\textbf{Type:} \colorbox{lavender}{Unable to parse long context tables}
\newline\newline
\textbf{Question:} “For clinical trial \texttt{NCT00720512}, among the more effective regimen ingredients, identify which ingredient starts with the letter ‘I’. Then, for this ingredient last approved up to December 2024, provide its patent expiration date (overall FDA approval only). Return only \texttt{YYYY}.”
& The agent attempted to extract patent‑expiry data from a multi‑page Orange Book PDF containing dense tables. Because it did not robustly parse the table structure, it surfaced only partial approval milestones (e.g., accelerated/full approvals for irinotecan) and never captured the patent‑expiration year requested, leading to an incomplete answer.\\
\bottomrule
\end{tabular}
\end{table}

\subsection{Computer Use Agent Results}
% \begin{figure}[H]
%   \centering
%   \begin{minipage}[c]{0.48\textwidth}
    
% Figure~\ref{fig:workflow} presents performance of Anthropic’s Computer Use agent (Sonnet 3.7) on a diagnostic subset of MedBrowseComp. This subset focuses on simpler tasks—each requiring no more than two browser actions—designed to isolate the agent’s ability to retrieve and extract specific biomedical fields with minimal navigational complexity. All tasks are grounded in HemOnc.org and span four clinically relevant question types: Clinical Trial IDs, Start Dates, Second Authors, and PubMed IDs (PMIDs).

%   \end{minipage}\hfill
%   %--- text on the left -----------------------------------
%   \begin{minipage}[c]{0.5\textwidth}
%       \centering
%         \includegraphics[width=\linewidth]{images/cua-performance.png}
%         \captionof{figure}{Anthropic Computer‑Use performance across tasks in Anthropic’s environment/sandbox among all our easier tasks grounded in hemonc}
%         \label{fig:workflow}
%   \end{minipage}
% \end{figure}
\begin{figure}[H]
  \centering
  % ---------- left column: explanatory text ----------
  \begin{minipage}[c]{0.5\textwidth}
Table~\ref{tab:cua} presents performance of Anthropic's Computer-Use
agent (Sonnet 3.7) on a diagnostic subset of \textsc{MedBrowseComp}.  This
subset focuses on simpler tasks comparing to the deepresearch partion—each requiring no more than two browser
actions—designed to isolate the agent's ability to retrieve and extract
specific biomedical fields with minimal navigational complexity.  All tasks
are grounded in \href{https://hemonc.org}{HemOnc.org} and span four clinically
relevant question types: \emph{Clinical Trial IDs}, \emph{Start Dates},
\emph{Second Authors}, and \emph{PubMed IDs (PMIDs)}.
  \end{minipage}\hfill
  % ---------- right column: performance table ----------
  \begin{minipage}[c]{0.45\textwidth}
    \centering
    \begin{tabular}{l c}
      \toprule
      \textbf{Extraction Task} & \textbf{Accuracy (\%)} \\
      \midrule
      Clinical Trial IDs & 33.88 \\
      Start Date         & 30.58 \\
      Second Author      & 36.36 \\
      PMIDs              & 11.57 \\
      \bottomrule
    \end{tabular}
    \captionof{table}{Anthropic Computer-Use performance on
    four HemOnc information-extraction tasks in the sandbox environment.}
    \label{tab:cua}
  \end{minipage}
\end{figure}

The agent performs best on finding the Second Author of the linked PubMed paper and Trial ID extraction, relatively low-hop tasks that benefit from predictable formatting and shallow navigation. In addition, this information is mostly be found in close proximity on the same page, which likely explains their similar results. Start Date questions require navigating to ClinicalTrials.gov and correctly identifying structured metadata. Errors in these settings often stem from greedy field selection: trajectory analysis shows the agent tends to extract the first date it encounters—even if unrelated to trial start—rather than locating the dedicated “Study Start” field. 

As we take one step further beyond Hemonc, asking for the matching PMID - despite appearing structurally simple - results in the weakest performance. Interestingly, this is not due to common visual brittleness or interface failures. Instead, the agent often returns a valid PMID for a plausible but incorrect paper. This suggests semantic confusion rather than surface-level noise. You can see a detailed common error modes sorted with explanations in the following Table \ref{tab:cua-common-errors}. 

\begin{table}[H]
\centering
\caption{Examples of common errors for Sonnet 3.7 on our CUA tasks. We provide three examples, each with attributed common error modes and a detailed explanation screened by the author PM.}

\begin{tabular}
{p{0.45\textwidth}p{0.5\textwidth}}
\toprule
\textbf{Error type \& question (paraphrased)} & \textbf{Explanation} \\
\midrule
\textbf{Type:} \colorbox{lavender}{Greedy attribute extraction} 
\newline\newline
\textbf{Question:} Give the start date (\texttt{YYYY‑MM}) for the trial Gemcitabin ± Cisplatin in cholangiocarcinoma. &
Agent located the ABC‑02 trial but copied “September 2006” from the paper’s timeline instead of scrolling down in search of the correct and meaningful start date. The registry lists 2005‑05; using a narrative source instead of the structured field produced a one‑year error. \\
\midrule
\textbf{Type:} \colorbox{lavender}{Overlooking information }
\newline\newline
\textbf{Questions:} Which clinical‑trial ID best describes Rituximab monotherapy compared to Ibrutinib monotherapy when used to treat Chronic lymphocytic leukemia. &
The model searched and correctly opened the relevant URL (\texttt{NCT01234311}). It scrolled through the page but stated it could not find an NCT ID despite it being present, so it continued the search and eventually retrieved \texttt{NCT01886872}. \\
\midrule

\textbf{Type:} \colorbox{lavender}{Aimless wondering}
\newline\newline
\textbf{Question:} “Give me the second author of the paper about the efficacy of Tranexamic acid monotherapy vs Placebo when used to treat Hereditary hemorrhagic telangiectasia. & The agent correctly opens the relevant NCT trial and attempts to find the associated paper online. However, the first result of the search is not the the actual study. Unable to identify the second author from that source, it explores the linked website for references to any related publications. It then opens the first paper it finds and returns the second author from that unrelated document.\\
\bottomrule
\label{tab:cua-common-errors}
\end{tabular}
\end{table}
\section{Conclusion and Future Work}Our work has certain \textbf{limitations}:

\textbf{System selection:}  We evaluated only agents with publicly programmable APIs.  As a result, promising but closed systems such as OpenAI \textit{Operator} and Google’s \textit{Project Mariner} were not evaluated. Other systems like Bytedance’s \textit{UITars} were not tested due to time\ and computational constraints \cite{qin2025ui}. However, it is not clear that their inclusion would shift the leaderboard substantially given their performance compared to Claude CUA on other benchmarks.

\textbf{Human supervision:}  All answers were machine‑judged with a lightly‑audited LLM rubric. However, a subset of answers were human-verified with good agreement compared to the LLM-as-judge.

\textbf{Compute and subscription cost:}  Building the benchmark and running baselines is non‑trivial and our experiments cost a total of \(\$3,690\): \(\$320\) on Perplexity (of which \(\$200\) was Deep‑Research API calls), \(\$450\) on Gemini 2.5 pro, and about \(\$2{,}500\) on Claude Sonnet 3.7 CUA, \(\$200\) for a 1 month ChatGPT Pro subscription, \(\$20\) for advanced reasoning, and \(\$200\) for GPT‑4.1 mini judging. 

\textbf{Real‑world validation:}  Finally, we did not place answers in front of clinicians , and the questions in MedBrowseComp only cover a small portion of the enormous medical knowledge base used for real-world clinical decision-making. However, expert-curated, verifiable knowledge is not available for the entire medical domain. Our focused benchmark enables deeper study of deep research and computer use capabilities and still reveals important gaps and future directions to inform the field.

\subsection{Future Work}

\textbf{Broader task suite:}  Expand beyond single‑field extraction to multi‑paragraph justification, guideline concordance, and financial/regulatory trend analysis-all of which require deeper reasoning.

\textbf{Tool‑augmented agents.}  Test whether lightweight adapters—PDF parsers, table detectors, ClinicalTrials.gov wrappers—allow computer use agents.

\textbf{Inclusion of closed systems.}  Collaborate with companies to benchmark Operator, Mariner, and UITars under identical prompts, closing the current comparability gap.

\textbf{Test system in AI-IDE environment}  We have not yet placed agents in an AI-IDE sandbox (e.g., Cursor or Windsurf) where they must draft, debug, and reuse their own helper scripts over our horizontal tasks. Such an environment would expose whether a model can sustain state, recover from bugs, and refactor code as requirements evolve.

\textbf{Study agents with a human-in-the-loop and in real-world settings}  Future work comparing agentic system vs human performance on MedBrowseComp, and studying top-performing agentic systems in real clinical settings, will be essential to understand the optimal implementation and clinical-translaational role of the benchmark and agentic systems.

\subsection{Conclusion}
We introduced \textsc{MedBrowseComp}, the largest verifiable benchmark that evaluates deep research and computer use agents in the medical field. The tasks presented in \textsc{MedBrowseComp} are not contrived for difficulty alone; each question mirrors a clinically meaningful information‑seeking scenario. Experiments reveal a clear capability gap: retrieval augmented text pipelines answer nearly half as many questions as their deep research counterparts, yet no system, including computer use agents, exceeds 40\% accuracy among questions that require more than a single hop.

We find that while GUI‑centric agents such as Anthropic’s \textit{Computer‑Use} demo can control browsers and desktops end‑to‑end, they tend to under-perform compared to deep researchers driven by APIs on our benchmark. Every GUI action produces roughly two screenshots, inflating the context window to \(\sim\!200{,}000\) tokens—compared with only \(\sim\!2{,}000\) tokens for an equivalent text‑only workflow. Deepresearch pipelines, by contrast, reach structured endpoints in a single call, eliminating the multi‑step visual loop and reducing both latency and cost. 
\textsc{MedBrowseComp} offers a realistic yet challenging test bed for future deep‑research systems and supplies a focused subset that remains demanding—though not prohibitive—for computer‑use agents. We hope it will accelerate progress toward efficient, high‑accuracy medical question answering across modalities.

\FloatBarrier   % keeps floats before refs if desired

% --------------------------- References -----------------------------------
\bibliographystyle{unsrtnat}  % numeric, ordered by citation
\bibliography{cross_care}

% --------------------------- Appendix & checklist -------------------------
\clearpage
\appendix
\newpage
% \todo{use paragraphs instead of paragraph}
\paragraph{Contribution}

\paragraph{Shan Chen} and \textbf{Pedro Moreira} contributed equally to this work. They developed the main theoretical framework, performed the experiments, and led the writing of the manuscript.

\textbf{Yuxin Xiao} was instrumental in experimental design and data collection.

\textbf{Sam Schmidgall}, \textbf{Jeremy Warner}, and \textbf{Hugo Aerts} were key during the dataset curations.

\textbf{Thomas Hartvigsen} and \textbf{Jack Gallifant} provided supervision, mentorship of junior team member, and was instrumental in the strategic direction of the research.

\textbf{Danielle S Bitterman}, the corresponding author, oversaw the entire project, coordinated the interdisciplinary team, and secured funding. She mentored junior team members and ensured the final approval of the version to be published.

\paragraph{Acknowledgement}
The authors thank Google Cloud for computing, API costs for testing. 

The authors also acknowledge financial support from Google PhD Fellowship (SC), Fullbright Scholarship and Erasmus Mundus JM Scholarship (PM), the Woods Foundation (DB, HA), NIH (NIH-USA U54CA274516-01A1 (SC, JG, HA, DB), NIH-USA R01CA294033 (DB, SC, JG) and the American Cancer Society
and American Society for Radiation Oncology, ASTRO-CSDG-24-1244514-01-CTPS Grant (DB). NIH-USA U24CA194354 (HA), NIH-USA U01CA190234 (HA), NIH-USA U01CA209414 (HA), and NIH-USA R35CA22052 (HA), and the European Union - European Research Council (HA: 866504) 

\paragraph{Ethics} \textsc{MedBrowseComp} is built solely from publicly available, non-identifiable sources, so no protected health information is exposed. Its results are intended for research benchmarking and must not be interpreted as clinical performance guarantees; any real-world deployment requires qualified human oversight. We acknowledge potential corpus-level biases and plan bias audits and broader source diversification in future releases.

\newpage
\section{Appendix}

\subsection{Detailed Benchmark Setting} \label{bench_details}
\begin{table}[htbp]
\centering
\resizebox{\textwidth}{!}{%
\begin{tabular}{l l l l l}
\toprule
\textbf{Model} & \textbf{Mode} & \textbf{Test Time} & \textbf{Version} & \textbf{Verification Method} \\
\midrule
\multicolumn{5}{c}{\textbf{MedBrowseComp 50}} \\
\midrule
O3 & search & April 29th, 2025 & Pro subscription & llm + human \\
O3 & deepresearch & April 29 -- May 1st, 2025 & Pro subscription & llm + human \\
Gemini2.5pro & search & May 1st, 2025 & api - 03/25 & llm + human \\
Gemini2.5pro & deepresearch & April 30 -- May 1st, 2025 & One subscription & llm + human \\
Sonar pro & search & April 28th, 2025 & api & llm + human \\
Perplexity & deepresearch & April 28th, 2025 & api & llm + human \\
Sonnet 3.7 & CUA & May 6th, 2025 & api - vertex 20250219 & llm + human \\
\midrule
\multicolumn{5}{c}{\textbf{MedBrowseComp 605}} \\
\midrule
Gemini2.5pro & search & May 2nd, 2025 & api - 03/25 & llm \\
GPT-4.1 & search-high & May 2nd, 2025 & api - gpt-4.1-mini-2025-04-14 & llm \\
Gemini2flash & search & May 2nd, 2025 & api & llm \\
Gemini2.5pro & search & May 1st, 2025 & api - 03/25 & llm \\
Sonar pro & search & May 2nd, 2025 & api & llm \\
\midrule
\multicolumn{5}{c}{\textbf{MedBrowseComp CUA}} \\
\midrule
Sonnet 3.7 & CUA & May 6th, 2025 & api - vertex 20250219 & llm \\
\bottomrule
\end{tabular}%
}
\caption{Overview of test runs for MedBrowseComp benchmarks.}
\label{tab:MedBrowseComp_benchmark_summary}
\end{table}

\newpage
\subsection{LLM as judge Prompts}
We mostly use the same additional instruction as Humanity’s Last Exam \cite{phan2025humanitysexam}. However we did some modification due to our need. First we used Openai structure output pipeline accompany with system prompt of \textit{You are an impartial judge evaluating an AI response based on provided criteria. Respond ONLY with a valid JSON object matching the requested structure.} 

We also did not include the confidence estimation as we do not have API access to many of the models.

\subsection*{Pydantic Schema for Structured Output}

\begin{framed}
\begin{lstlisting}
class JudgeOutput(BaseModel):
    extracted_final_answer: str = Field(
        description="The final exact answer extracted from the [response]. Put 'None' if no exact answer found."
    )
    reasoning: str = Field(
        description="Explanation of correctness based ONLY on comparing extracted_final_answer and correct_answer."
    )
    correct: str = Field(
        description="Must be 'yes' or 'no'."
    )
\end{lstlisting}
\end{framed}

\subsection*{Grading Prompt}
\begin{framed}
\ttfamily
JUDGE\_PROMPT = """Judge whether the following [response] to [question] is correct or not based on the\\
precise and unambiguous [correct\_answer] below.\\
\texttt{[question]}: \{question\}\\
\texttt{[response]}: \{response\}\\
Your judgement must be in the format and criteria specified below:\\
extracted\_final\_answer: The final exact answer extracted from the [response]. Put the extracted answer\\
as 'None' if there is no exact, final answer to extract from the response.\\
\texttt{[correct\_answer]}: \{correct\_answer\}\\
reasoning: Explain why the extracted\_final\_answer is correct or incorrect based on [correct\_answer],\\
focusing only on if there are meaningful differences between [correct\_answer] and the\\
extracted\_final\_answer. Do not comment on any background to the problem, do not attempt to solve\\
the problem, do not argue for any answer different than [correct\_answer], focus only on whether the\\
answers match.\\
correct: Answer 'yes' if extracted\_final\_answer matches the \texttt{[correct\_answer]} given above, or is within\\
a small margin of error for numerical problems. Answer 'no' otherwise, i.e. if there is\\
any inconsistency, ambiguity, non-equivalency, or if the extracted answer is incorrect."""
\end{framed}

\newpage
\subsection{Computer Use Agent Prompt}
\label{CUA-prompt}
The prompt remains largely identical to the original version provided by Anthropic, with a single minor addition highlighted in gray.

A concise instruction reinforcing that the agent should act independently and refrain from requesting clarification or human assistance during execution, promoting model autonomy.

\begin{framed}
\ttfamily
\textless SYSTEM\_CAPABILITY\textgreater

\begin{itemize}
  \item You are utilising an Ubuntu virtual machine using \verb|{platform.machine()}| architecture with internet access.
  \item You can feel free to install Ubuntu applications with your bash tool. Use \texttt{curl} instead of \texttt{wget}.
  \item To open Firefox, please just click on the Firefox icon. Note, \texttt{firefox-esr} is what is installed on your system.
  \item Using bash tool you can start GUI applications, but you need to set \verb|export DISPLAY=:1| and use a subshell. For example, \verb|(DISPLAY=:1 xterm &)|. GUI apps run with bash tool will appear within your desktop environment, but they may take some time to appear. Take a screenshot to confirm it did.
  \item When using your bash tool with commands that are expected to output very large quantities of text, redirect into a tmp file and use \texttt{str\_replace\_editor} or \texttt{grep -n -B <lines before> -A <lines after> <query> <filename>} to confirm output.
  \item When viewing a page it can be helpful to zoom out so that you can see everything on the page. Either that, or make sure you scroll down to see everything before deciding something isn't available.
  \item When using your computer function calls, they take a while to run and send back to you. Where possible/feasible, try to chain multiple of these calls all into one function calls request.
  \item The current date is \verb|{datetime.today().strftime('%A, %B %-d, %Y')}|.
\end{itemize}

\textless /SYSTEM\_CAPABILITY\textgreater

\vspace{1em}
\textless IMPORTANT\textgreater

\begin{itemize}
  \item \colorbox{lightgray}{\parbox{\dimexpr\linewidth-2\fboxsep}{\textbf{Never ask the user for help or to clarify. You are the assistant and you should be able to figure out what to do.}}}

  \item When using Firefox, if a startup wizard appears, \textbf{IGNORE IT}. Do not even click ``skip this step.'' Instead, click on the address bar where it says ``Search or enter address,'' and enter the appropriate search term or URL there.
  \item If the item you are looking at is a PDF, and after taking a single screenshot of the PDF it seems that you want to read the entire document instead of trying to continue to read the PDF from your screenshots + navigation, determine the URL, use \texttt{curl} to download the PDF, install and use \texttt{pdftotext} to convert it to a text file, and then read that text file directly with your \texttt{StrReplaceEditTool}.
\end{itemize}

\textless /IMPORTANT\textgreater
\end{framed}

\newpage
\subsection{Deep Research Agent Results} \label{deepresearch_results}

\begin{table}[ht]
\centering
\caption{Detailed Performance of Frontier Systems on MedBrowseComp 50 - For this subset, we selected where the questions cannot be answered by NA}
\label{tab:frontier-systems-expanded}
\begin{tabular}{l*{7}{c}}
\toprule
\multirow{2}{*}{\textbf{Question Depth}} 
  & \multicolumn{2}{c}{\textbf{O3}} 
  & \multicolumn{2}{c}{\textbf{Gemini2.5pro}} 
  & \multicolumn{2}{c}{\textbf{Perplexity}} 
  & \multirow{2}{*}{\textbf{Claude-CUA}} \\
\cmidrule(lr){2-3} \cmidrule(lr){4-5} \cmidrule(lr){6-7}
& search & deep & search & deep & search & deep & \\
\midrule

\multirow{2}{*}{1-hop (n=10)} 
& \makecell{10/10 \\ \scriptsize(\textbf{100.0\%})} & \makecell{10/10 \\ \scriptsize(\textbf{100.0\%})}
& \makecell{9/10 \\ \scriptsize(90.0\%)} & \makecell{10/10 \\ \scriptsize(\textbf{100.0\%})}
& \makecell{8/10 \\ \scriptsize(80.0\%)} & \makecell{10/10 \\ \scriptsize(\textbf{100.0\%})}
& \makecell{10/10 \\ \scriptsize(\textbf{100.0\%})} \\

\addlinespace

\multirow{2}{*}{2-hop (n=10)} 
& \makecell{5/10 \\ \scriptsize(50.0\%)} & \makecell{6.5/10 \\ \scriptsize(65.0\%)}
& \makecell{4/10 \\ \scriptsize(40.0\%)} & \makecell{\textbf{8}/10 \\ \scriptsize(\textbf{80.0\%})}
& \makecell{2/10 \\ \scriptsize(20.0\%)} & \makecell{5/10 \\ \scriptsize(50.0\%)}
& \makecell{4/10 \\ \scriptsize(40.0\%)} \\

\addlinespace

\multirow{2}{*}{3-hop (n=10)} 
& \makecell{1/10 \\ \scriptsize(10.0\%)} & \makecell{3/10 \\ \scriptsize(30.0\%)}
& \makecell{1/10 \\ \scriptsize(10.0\%)} & \makecell{\textbf{3.5}/10 \\ \scriptsize(\textbf{35.0\%})}
& \makecell{0/10 \\ \scriptsize(0.0\%)} & \makecell{3/10 \\ \scriptsize(30.0\%)}
& \makecell{2/10 \\ \scriptsize(20.0\%)} \\

\addlinespace

\multirow{2}{*}{4-hop (n=10)} 
& \makecell{3/10 \\ \scriptsize(30.0\%)} & \makecell{\textbf{5}/10 \\ \scriptsize(\textbf{50.0\%})}
& \makecell{0/10 \\ \scriptsize(0.0\%)} & \makecell{3/10 \\ \scriptsize(30.0\%)}
& \makecell{0/10 \\ \scriptsize(0.0\%)} & \makecell{2/10 \\ \scriptsize(20.0\%)}
& \makecell{1/10 \\ \scriptsize(10.0\%)} \\

\addlinespace

\multirow{2}{*}{5-hop (n=10)} 
& \makecell{0/10 \\ \scriptsize(0.0\%)} & \makecell{\textbf{1}/10 \\ \scriptsize(\textbf{10.0\%})}
& \makecell{0/10 \\ \scriptsize(0.0\%)} & \makecell{0/10 \\ \scriptsize(0.0\%)}
& \makecell{0/10 \\ \scriptsize(0.0\%)} & \makecell{0/10 \\ \scriptsize(0.0\%)}
& \makecell{1/10 \\ \scriptsize(10.0\%)} \\

\midrule

\multirow{2}{*}{\textbf{Total (n=50)}} 
& \makecell{19/50 \\ \scriptsize(38.0\%)} & \makecell{\textbf{25.5}/50 \\ \scriptsize(\textbf{51.0\%})}
& \makecell{14/50 \\ \scriptsize(28.0\%)} & \makecell{24.5/50 \\ \scriptsize(49.0\%)}
& \makecell{10/50 \\ \scriptsize(20.0\%)} & \makecell{20/50 \\ \scriptsize(40.0\%)}
& \makecell{18/50 \\ \scriptsize(36.0\%)} \\

\bottomrule
\end{tabular}
\end{table}

The benchmark we’ve created is tough as you can see from MedBrowseComp 50 and more detailed results on MedBrowseComp 605 on the following page. It’s designed to push models beyond just recognizing patterns or guessing from context. Instead, it asks them to follow a chain of reasoning across multiple steps (or “hops”) through a medical knowledge base. And when you strip away the ability to say “Not applicable” or avoid answering (what we call REAL accuracy), the results show just how hard this is. Even the strongest model, Gemini 2.5 Pro (search) , only gets about 24.5\% of the answers right under these strict conditions. That might not sound like much, but it still makes it the clear leader in this group.

What’s especially telling is how performance drops off with each additional hop. For example, Gemini 2.5 Pro does well on 1-hop questions (76\%), where the answer is often directly stated. But by the time you get to 4-hop or 5-hop questions — where the model has to link together several pieces of information in sequence — even it struggles. On REAL accuracy for 4-hop questions, it only gets 5.1\% , and for 5-hop, it’s essentially zero. This shows that while models may look good on simple tasks, chaining together multiple steps of reasoning is still a big challenge.

In short, we hope this benchmark doesn’t let models take shortcuts. We want to force them to dig into real medical knowledge and reason carefully. And based on these results, there’s still a long way to go before we can fully trust AI systems to handle complex, multi-step medical reasoning without supervision.

\newpage
\begin{table}[H]
\centering
\caption{Detailed Performance of Models on MedBrowseComp 605 | Note that SonarPro-param is blank here due to the lack of non-search options from perplexity.}
\label{tab:na-acc}
\begin{tabular}{l*{9}{c}}
\toprule
\multirow{2}{*}{\textbf{Question Depth}} 
  & \multicolumn{2}{c}{\textbf{GPT-4.1}} 
  & \multicolumn{2}{c}{\textbf{SonarPro}} 
  & \multicolumn{2}{c}{\textbf{Gemini2Flash}} 
  & \multicolumn{2}{c}{\textbf{GeminiPro}} \\
\cmidrule(lr){2-3} \cmidrule(lr){4-5} \cmidrule(lr){6-7} \cmidrule(lr){8-9}
& param & search & param & search & param & search & param & search \\
\midrule

\multirow{2}{*}{1-hop (n=121)} 
& \makecell{24/121 \\ \scriptsize(19.8\%)} 
& \makecell{19/121 \\ \scriptsize(15.7\%)} 
& -- & \makecell{63/121 \\ \scriptsize(52.1\%)} 
& \makecell{26/121 \\ \scriptsize(21.5\%)} & \makecell{67/121 \\ \scriptsize(55.4\%)} 
& \makecell{10/121 \\ \scriptsize(8.3\%)} & \makecell{\textbf{92}/121 \\ \scriptsize(\textbf{76.0\%})} \\

\addlinespace

\multirow{2}{*}{2-hop (n=121)} 
& \makecell{5/121 \\ \scriptsize(4.1\%)} 
& \makecell{5/121 \\ \scriptsize(4.1\%)} 
& -- & \makecell{8/121 \\ \scriptsize(6.6\%)} 
& \makecell{2/121 \\ \scriptsize(1.7\%)} & \makecell{7/121 \\ \scriptsize(5.8\%)} 
& \makecell{4/121 \\ \scriptsize(3.3\%)} & \makecell{\textbf{13}/121 \\ \scriptsize(\textbf{10.7\%})} \\

\addlinespace

\multirow{2}{*}{3-hop (n=121)} 
& \makecell{1/121 \\ \scriptsize(0.8\%)} 
& \makecell{1/121 \\ \scriptsize(0.8\%)} 
& -- & \makecell{2/121 \\ \scriptsize(1.7\%)} 
& \makecell{2/121 \\ \scriptsize(1.7\%)} & \makecell{1/121 \\ \scriptsize(0.8\%)} 
& \makecell{9/121 \\ \scriptsize(7.4\%)} & \makecell{\textbf{4}/121 \\ \scriptsize(\textbf{3.3\%})} \\

\addlinespace

\multirow{2}{*}{4-hop (n=121)} 
& \makecell{60/121 \\ \scriptsize(49.6\%)} 
& \makecell{42/121 \\ \scriptsize(34.7\%)} 
& -- & \makecell{70/121 \\ \scriptsize(57.9\%)} 
& \makecell{60/121 \\ \scriptsize(49.6\%)} & \makecell{48/121 \\ \scriptsize(39.7\%)} 
& \makecell{39/121 \\ \scriptsize(32.2\%)} & \makecell{\textbf{49}/121 \\ \scriptsize(\textbf{40.5\%})} \\

\addlinespace

\multirow{2}{*}{5-hop (n=121)} 
& \makecell{15/121 \\ \scriptsize(12.4\%)} 
& \makecell{12/121 \\ \scriptsize(9.9\%)} 
& -- & \makecell{15/121 \\ \scriptsize(12.4\%)} 
& \makecell{23/121 \\ \scriptsize(19.0\%)} & \makecell{15/121 \\ \scriptsize(12.4\%)} 
& \makecell{18/121 \\ \scriptsize(14.9\%)} & \makecell{\textbf{24}/121 \\ \scriptsize(\textbf{19.8\%})} \\

\midrule

\multirow{2}{*}{\textbf{Total (n=605)}} 
& \makecell{105/605 \\ \scriptsize(17.3\%)} 
& \makecell{80/605 \\ \scriptsize(13.2\%)} 
& -- & \makecell{158/605 \\ \scriptsize(26.1\%)} 
& \makecell{113/605 \\ \scriptsize(18.7\%)} & \makecell{138/605 \\ \scriptsize(22.8\%)} 
& \makecell{80/605 \\ \scriptsize(13.2\%)} & \makecell{\textbf{182}/605 \\ \scriptsize(\textbf{30.1\%})} \\

\bottomrule
\end{tabular}
\end{table}

\begin{table}[H]
\centering
\caption{REAL Accuracy for MedBrowseComp605 (Excluding NA-like Correct Answers): Models are evaluated only on questions where the correct answer is applicable. NA-like responses (e.g., "Not applicable") are excluded from scoring.}
\label{tab:real-acc}
\begin{tabular}{l*{9}{c}}
\toprule
\multirow{2}{*}{\textbf{Question Depth}} 
  & \multicolumn{2}{c}{\textbf{GPT-4.1}} 
  & \multicolumn{2}{c}{\textbf{SonarPro}} 
  & \multicolumn{2}{c}{\textbf{Gemini2Flash}} 
  & \multicolumn{2}{c}{\textbf{GeminiPro}} \\
\cmidrule(lr){2-3} \cmidrule(lr){4-5} \cmidrule(lr){6-7} \cmidrule(lr){8-9}
& param & search & param & search & param & search & param & search \\
\midrule

\multirow{2}{*}{1-hop (n=121)} 
& \makecell{24/121 \\ \scriptsize(19.8\%)} 
& \makecell{19/121 \\ \scriptsize(15.7\%)} 
& -- & \makecell{63/121 \\ \scriptsize(52.1\%)} 
& \makecell{26/121 \\ \scriptsize(21.5\%)} & \makecell{67/121 \\ \scriptsize(55.4\%)} 
& \makecell{10/121 \\ \scriptsize(8.3\%)} & \makecell{\textbf{92}/121 \\ \scriptsize(\textbf{76.0\%})} \\

\addlinespace

\multirow{2}{*}{2-hop (n=121)} 
& \makecell{5/121 \\ \scriptsize(4.1\%)} 
& \makecell{5/121 \\ \scriptsize(4.1\%)} 
& -- & \makecell{8/121 \\ \scriptsize(6.6\%)} 
& \makecell{2/121 \\ \scriptsize(1.7\%)} & \makecell{7/121 \\ \scriptsize(5.8\%)} 
& \makecell{4/121 \\ \scriptsize(3.3\%)} & \makecell{\textbf{13}/121 \\ \scriptsize(\textbf{10.7\%})} \\

\addlinespace

\multirow{2}{*}{3-hop (n=121)} 
& \makecell{1/121 \\ \scriptsize(0.8\%)} 
& \makecell{1/121 \\ \scriptsize(0.8\%)} 
& -- & \makecell{2/121 \\ \scriptsize(1.7\%)} 
& \makecell{2/121 \\ \scriptsize(1.7\%)} & \makecell{1/121 \\ \scriptsize(0.8\%)} 
& \makecell{9/121 \\ \scriptsize(7.4\%)} & \makecell{\textbf{4}/121 \\ \scriptsize(\textbf{3.3\%})} \\

\addlinespace

\multirow{2}{*}{4-hop (n=39)} 
& \makecell{0/39 \\ \scriptsize(0.0\%)} 
& \makecell{0/39 \\ \scriptsize(0.0\%)} 
& -- & \makecell{0/39 \\ \scriptsize(0.0\%)} 
& \makecell{0/39 \\ \scriptsize(0.0\%)} & \makecell{0/39 \\ \scriptsize(0.0\%)} 
& \makecell{0/39 \\ \scriptsize(0.0\%)} & \makecell{\textbf{2}/39 \\ \scriptsize(\textbf{5.1\%})} \\

\addlinespace

\multirow{2}{*}{5-hop (n=51)} 
& \makecell{0/51 \\ \scriptsize(0.0\%)} 
& \makecell{0/51 \\ \scriptsize(0.0\%)} 
& -- & \makecell{1/51 \\ \scriptsize(2.0\%)} 
& \makecell{1/51 \\ \scriptsize(2.0\%)} & \makecell{0/51 \\ \scriptsize(0.0\%)} 
& \makecell{0/51 \\ \scriptsize(0.0\%)} & \makecell{\textbf{0}/51 \\ \scriptsize(\textbf{0.0\%})} \\

\midrule

\multirow{2}{*}{\textbf{Total (n=453)}} 
& \makecell{30/453 \\ \scriptsize(6.6\%)} 
& \makecell{25/453 \\ \scriptsize(5.5\%)} 
& -- & \makecell{74/453 \\ \scriptsize(16.3\%)} 
& \makecell{31/453 \\ \scriptsize(6.8\%)} & \makecell{75/453 \\ \scriptsize(16.6\%)} 
& \makecell{23/453 \\ \scriptsize(5.1\%)} & \makecell{\textbf{111}/453 \\ \scriptsize(\textbf{24.5\%})} \\

\bottomrule
\end{tabular}
\end{table}

\newpage
\subsection{Evaluation from Optimal Start Page}

To evaluate the impact of structured entrypoints on Computer Use Agent performance, we re-ran the benchmark using the same Claude 3.7 Sonnet system but initialized each task from the \href{https://hemonc.org}{HemOnc.org homepage}. This strategy avoids ambiguity introduced by external search engines and leverages HemOnc.org's curated structure to simplify task execution.

\begin{table}[H]
    \centering
    \caption{Accuracy (\%) of Claude 3.7 Computer-Use Agent on structured extraction tasks, with and without initialization from HemOnc.org homepage. Improvements are shown in bold.}
    \label{tab:CUA-hemonc-comparison}
    \begin{tabular}{lcc}
        \toprule
        \textbf{Extraction Task} & \textbf{With HemOnc Start} & \textbf{No Defined Start} \\
        \midrule
        PMIDs          & \textbf{42.98} & 11.57 \\
        Second Author  & \textbf{46.28} & 36.36 \\
        NCT            & \textbf{39.67} & 33.88 \\
        Start Date     & \textbf{31.40} & 30.58 \\
        \bottomrule
    \end{tabular}
\end{table}

Compared to the results reported in Table~\ref{tab:MedBrowseComp_benchmark_summary}, these scores show notable improvement, particularly on PMID extraction, which increased from 11.57\% to 42.98\%. By starting from a high-quality structured resource, the agent appears less likely to encounter ambiguous links, irrelevant documents, or noisy intermediate pages.

We hypothesize several contributing factors:
\begin{itemize}
\item \textbf{Reduced ambiguity at the root}: Starting on HemOnc.org anchors the agent to a semantically dense, domain-verified hub. This prevents early misrouting or unnecessary detours caused by irrelevant or overly generic search results.
\item \textbf{Fewer hops, higher precision}: Structured navigation from HemOnc often yields answers in one or two clicks. This minimizes the risk of cumulative tool-call errors, selector mismatches, or hallucinated document interpretations.
\item \textbf{Improved alignment with human workflows}: Clinicians often use HemOnc as a first step for navigating oncology evidence. Our findings suggest that model workflows, like human ones, benefit from strong priors.
\end{itemize}

These findings suggest that GUI-agent accuracy can benefit significantly from constrained yet clinically meaningful entrypoints like HemOnc.org. Future work should explore formal initialization policies as part of web-agent pipelines.

\clearpage
\section*{NeurIPS Paper Checklist}

\begin{enumerate}

\item {\bf Claims}
    \item[] Question: Do the main claims made in the abstract and introduction accurately reflect the paper's contributions and scope?
    \item[] Answer: \answerYes{}
    \item[] Justification: Contributions are clearly enumerated at the end of the introduction, highlighting results and resources that can be found within the manuscript. 
    \item[] Guidelines:
    \begin{itemize}
        \item The answer NA means that the abstract and introduction do not include the claims made in the paper.
        \item The abstract and/or introduction should clearly state the claims made, including the contributions made in the paper and important assumptions and limitations. A No or NA answer to this question will not be perceived well by the reviewers. 
        \item The claims made should match theoretical and experimental results, and reflect how much the results can be expected to generalize to other settings. 
        \item It is fine to include aspirational goals as motivation as long as it is clear that these goals are not attained by the paper. 
    \end{itemize}

\item {\bf Limitations}
    \item[] Question: Does the paper discuss the limitations of the work performed by the authors?
    \item[] Answer: \answerYes{}
    \item[] Justification: A dedicated limitations section can be found at the end of the paper.  
    \item[] Guidelines:
    \begin{itemize}
        \item The answer NA means that the paper has no limitation while the answer No means that the paper has limitations, but those are not discussed in the paper. 
        \item A separate "Limitations" section in the paper clearly enumerates the key limitations of this paper.
        \item The paper should point out any strong assumptions and how robust the results are to violations of these assumptions (e.g., independence assumptions, noiseless settings, model well-specification, asymptotic approximations only holding locally). The authors should reflect on how these assumptions might be violated in practice and what the implications would be.
        \item The authors should reflect on the scope of the claims made, e.g., if the approach was only tested on a few datasets or with a few runs. In general, empirical results often depend on implicit assumptions, which should be articulated.
        \item The authors should reflect on the factors that influence the performance of the approach. For example, a facial recognition algorithm may perform poorly when image resolution is low or images are taken in low lighting. Or a speech-to-text system might not be used reliably to provide closed captions for online lectures because it fails to handle technical jargon.
        \item The authors should discuss the computational efficiency of the proposed algorithms and how they scale with dataset size.
        \item If applicable, the authors should discuss possible limitations of their approach to address problems of privacy and fairness.
        \item While the authors might fear that complete honesty about limitations might be used by reviewers as grounds for rejection, a worse outcome might be that reviewers discover limitations that aren't acknowledged in the paper. The authors should use their best judgment and recognize that individual actions in favor of transparency play an important role in developing norms that preserve the integrity of the community. Reviewers will be specifically instructed to not penalize honesty concerning limitations.
    \end{itemize}

\item {\bf Theory Assumptions and Proofs}
    \item[] Question: For each theoretical result, does the paper provide the full set of assumptions and a complete (and correct) proof?
    \item[] Answer: \answerNA{}.
    \item[] Justification: No theoretical results are presented in this piece. Any calculations have associated equations in-line and are referenced as such.
    \item[] Guidelines:
    \begin{itemize}
        \item The answer NA means that the paper does not include theoretical results. 
        \item All the theorems, formulas, and proofs in the paper should be numbered and cross-referenced.
        \item All assumptions should be clearly stated or referenced in the statement of any theorems.
        \item The proofs can either appear in the main paper or the supplemental material, but if they appear in the supplemental material, the authors are encouraged to provide a short proof sketch to provide intuition. 
        \item Inversely, any informal proof provided in the core of the paper should be complemented by formal proofs provided in appendix or supplemental material.
        \item Theorems and Lemmas that the proof relies upon should be properly referenced. 
    \end{itemize}

    \item {\bf Experimental Result Reproducibility}
    \item[] Question: Does the paper fully disclose all the information needed to reproduce the main experimental results of the paper to the extent that it affects the main claims and/or conclusions of the paper (regardless of whether the code and data are provided or not)?
    \item[] Answer: \answerYes{}
    \item[] Justification: All code is available in a public repository.  
    \item[] Guidelines:
    \begin{itemize}
        \item The answer NA means that the paper does not include experiments.
        \item If the paper includes experiments, a No answer to this question will not be perceived well by the reviewers: Making the paper reproducible is important, regardless of whether the code and data are provided or not.
        \item If the contribution is a dataset and/or model, the authors should describe the steps taken to make their results reproducible or verifiable. 
        \item Depending on the contribution, reproducibility can be accomplished in various ways. For example, if the contribution is a novel architecture, describing the architecture fully might suffice, or if the contribution is a specific model and empirical evaluation, it may be necessary to either make it possible for others to replicate the model with the same dataset, or provide access to the model. In general. releasing code and data is often one good way to accomplish this, but reproducibility can also be provided via detailed instructions for how to replicate the results, access to a hosted model (e.g., in the case of a large language model), releasing of a model checkpoint, or other means that are appropriate to the research performed.
        \item While NeurIPS does not require releasing code, the conference does require all submissions to provide some reasonable avenue for reproducibility, which may depend on the nature of the contribution. For example
        \begin{enumerate}
            \item If the contribution is primarily a new algorithm, the paper should make it clear how to reproduce that algorithm.
            \item If the contribution is primarily a new model architecture, the paper should describe the architecture clearly and fully.
            \item If the contribution is a new model (e.g., a large language model), then there should either be a way to access this model for reproducing the results or a way to reproduce the model (e.g., with an open-source dataset or instructions for how to construct the dataset).
            \item We recognize that reproducibility may be tricky in some cases, in which case authors are welcome to describe the particular way they provide for reproducibility. In the case of closed-source models, it may be that access to the model is limited in some way (e.g., to registered users), but it should be possible for other researchers to have some path to reproducing or verifying the results.
        \end{enumerate}
    \end{itemize}

\item {\bf Open access to data and code}
    \item[] Question: Does the paper provide open access to the data and code, with sufficient instructions to faithfully reproduce the main experimental results, as described in supplemental material?
    \item[] Answer: \answerYes{}
    \item[] Justification: A detailed README has been provided within each repository folder describing the steps required to reproduce or extend the current work. All final counts, outputs, and LLM as judge results are available for download on the public website. 
    \item[] Guidelines:
    \begin{itemize}
        \item The answer NA means that paper does not include experiments requiring code.
        \item Please see the NeurIPS code and data submission guidelines (\url{https://nips.cc/public/guides/CodeSubmissionPolicy}) for more details.
        \item While we encourage the release of code and data, we understand that this might not be possible, so “No” is an acceptable answer. Papers cannot be rejected simply for not including code, unless this is central to the contribution (e.g., for a new open-source benchmark).
        \item The instructions should contain the exact command and environment needed to run to reproduce the results. See the NeurIPS code and data submission guidelines (\url{https://nips.cc/public/guides/CodeSubmissionPolicy}) for more details.
        \item The authors should provide instructions on data access and preparation, including how to access the raw data, preprocessed data, intermediate data, and generated data, etc.
        \item The authors should provide scripts to reproduce all experimental results for the new proposed method and baselines. If only a subset of experiments are reproducible, they should state which ones are omitted from the script and why.
        \item At submission time, to preserve anonymity, the authors should release anonymized versions (if applicable).
        \item Providing as much information as possible in supplemental material (appended to the paper) is recommended, but including URLs to data and code is permitted.
    \end{itemize}

\item {\bf Experimental Setting/Details}
    \item[] Question: Does the paper specify all the training and test details (e.g., data splits, hyperparameters, how they were chosen, type of optimizer, etc.) necessary to understand the results?
    \item[] Answer:\answerYes{}
    \item[] Justification: While no training or tuning was conducted, we provided all our code, settings and outputs. 
    \item[] Guidelines:
    \begin{itemize}
        \item The answer NA means that the paper does not include experiments.
        \item The experimental setting should be presented in the core of the paper to a level of detail that is necessary to appreciate the results and make sense of them.
        \item The full details can be provided either with the code, in appendix, or as supplemental material.
    \end{itemize}

\item {\bf Experiment Statistical Significance}
    \item[] Question: Does the paper report error bars suitably and correctly defined or other appropriate information about the statistical significance of the experiments?
    \item[] Answer: \answerNA{} % Replace by \answerYes{}, \answerNo{}, or \answerNA{}.
    \item[] Justification: Given the nature of benchmarking and excessive costs plus we do not have API access to many of the services. All of the results we provided are pass@1. 
    
    \item[] Guidelines:
    \begin{itemize}
        \item The answer NA means that the paper does not include experiments.
        \item The authors should answer "Yes" if the results are accompanied by error bars, confidence intervals, or statistical significance tests, at least for the experiments that support the main claims of the paper.
        \item The factors of variability that the error bars are capturing should be clearly stated (for example, train/test split, initialization, random drawing of some parameter, or overall run with given experimental conditions).
        \item The method for calculating the error bars should be explained (closed form formula, call to a library function, bootstrap, etc.)
        \item The assumptions made should be given (e.g., Normally distributed errors).
        \item It should be clear whether the error bar is the standard deviation or the standard error of the mean.
        \item It is OK to report 1-sigma error bars, but one should state it. The authors should preferably report a 2-sigma error bar than state that they have a 96\% CI, if the hypothesis of Normality of errors is not verified.
        \item For asymmetric distributions, the authors should be careful not to show in tables or figures symmetric error bars that would yield results that are out of range (e.g. negative error rates).
        \item If error bars are reported in tables or plots, The authors should explain in the text how they were calculated and reference the corresponding figures or tables in the text.
    \end{itemize}

\item {\bf Experiments Compute Resources}
    \item[] Question: For each experiment, does the paper provide sufficient information on the computer resources (type of compute workers, memory, time of execution) needed to reproduce the experiments?
    \item[] Answer: \answerYes{}
    \item[] Justification: They are all included in our conclusion section.
    \item[] Guidelines:
    \begin{itemize}
        \item The answer NA means that the paper does not include experiments.
        \item The paper should indicate the type of compute workers CPU or GPU, internal cluster, or cloud provider, including relevant memory and storage.
        \item The paper should provide the amount of compute required for each of the individual experimental runs as well as estimate the total compute. 
        \item The paper should disclose whether the full research project required more compute than the experiments reported in the paper (e.g., preliminary or failed experiments that didn't make it into the paper). 
    \end{itemize}
    
\item {\bf Code Of Ethics}
    \item[] Question: Does the research conducted in the paper conform, in every respect, with the NeurIPS Code of Ethics \url{https://neurips.cc/public/EthicsGuidelines}?
    \item[] Answer: \answerYes{}
    \item[] Justification: The authors of this study have read, and confirm this study conforms with every aspect of the Code of Ethics.
    \item[] Guidelines:
    \begin{itemize}
        \item The answer NA means that the authors have not reviewed the NeurIPS Code of Ethics.
        \item If the authors answer No, they should explain the special circumstances that require a deviation from the Code of Ethics.
        \item The authors should make sure to preserve anonymity (e.g., if there is a special consideration due to laws or regulations in their jurisdiction).
    \end{itemize}

\item {\bf Broader Impacts}
    \item[] Question: Does the paper discuss both potential positive societal impacts and negative societal impacts of the work performed?
    \item[] Answer: \answerYes{} % Replace by \answerYes{}, \answerNo{}, or \answerNA{}.
    \item[] Justification: We do not think our paper holds many negative societal impacts. We did include an ethic section to discuss in Section 5.  And, we are eager to discuss and include if proper during the rebuttal. 
    \item[] Guidelines:
    \begin{itemize}
        \item The answer NA means that there is no societal impact of the work performed.
        \item If the authors answer NA or No, they should explain why their work has no societal impact or why the paper does not address societal impact.
        \item Examples of negative societal impacts include potential malicious or unintended uses (e.g., disinformation, generating fake profiles, surveillance), fairness considerations (e.g., deployment of technologies that could make decisions that unfairly impact specific groups), privacy considerations, and security considerations.
        \item The conference expects that many papers will be foundational research and not tied to particular applications, let alone deployments. However, if there is a direct path to any negative applications, the authors should point it out. For example, it is legitimate to point out that an improvement in the quality of generative models could be used to generate deepfakes for disinformation. On the other hand, it is not needed to point out that a generic algorithm for optimizing neural networks could enable people to train models that generate Deepfakes faster.
        \item The authors should consider possible harms that could arise when the technology is being used as intended and functioning correctly, harms that could arise when the technology is being used as intended but gives incorrect results, and harms following from (intentional or unintentional) misuse of the technology.
        \item If there are negative societal impacts, the authors could also discuss possible mitigation strategies (e.g., gated release of models, providing defenses in addition to attacks, mechanisms for monitoring misuse, mechanisms to monitor how a system learns from feedback over time, improving the efficiency and accessibility of ML).
    \end{itemize}
    
\item {\bf Safeguards}
    \item[] Question: Does the paper describe safeguards that have been put in place for responsible release of data or models that have a high risk for misuse (e.g., pretrained language models, image generators, or scraped datasets)?
    \item[] Answer: \answerYes{}{} % Replace by \answerYes{}, \answerNo{}, or \answerNA{}.
    \item[] Justification: All models and datasets utilized in this study are already publicly available. However, to prevent pre-training contamination in from scraping GitHub, we include an encoded version of our dataset publicly.
    \item[] Guidelines:
    \begin{itemize}
        \item The answer NA means that the paper poses no such risks.
        \item Released models that have a high risk for misuse or dual-use should be released with necessary safeguards to allow for controlled use of the model, for example by requiring that users adhere to usage guidelines or restrictions to access the model or implementing safety filters. 
        \item Datasets that have been scraped from the Internet could pose safety risks. The authors should describe how they avoided releasing unsafe images.
        \item We recognize that providing effective safeguards is challenging, and many papers do not require this, but we encourage authors to take this into account and make a best faith effort.
    \end{itemize}

\item {\bf Licenses for existing assets}
    \item[] Question: Are the creators or original owners of assets (e.g., code, data, models), used in the paper, properly credited and are the license and terms of use explicitly mentioned and properly respected?
    \item[] Answer: \answerYes{} % Replace by \answerYes{}, \answerNo{}, or \answerNA{}.
    \item[] Justification: All datasets are open access and comply with the copyright and terms of service under Apache 2.
    \item[] Guidelines:
    \begin{itemize}
        \item The answer NA means that the paper does not use existing assets.
        \item The authors should cite the original paper that produced the code package or dataset.
        \item The authors should state which version of the asset is used and, if possible, include a URL.
        \item The name of the license (e.g., CC-BY 4.0) should be included for each asset.
        \item For scraped data from a particular source (e.g., website), the copyright and terms of service of that source should be provided.
        \item If assets are released, the license, copyright information, and terms of use in the package should be provided. For popular datasets, \url{paperswithcode.com/datasets} has curated licenses for some datasets. Their licensing guide can help determine the license of a dataset.
        \item For existing datasets that are re-packaged, both the original license and the license of the derived asset (if it has changed) should be provided.
        \item If this information is not available online, the authors are encouraged to reach out to the asset's creators.
    \end{itemize}

\item {\bf New Assets}
    \item[] Question: Are new assets introduced in the paper well documented and is the documentation provided alongside the assets?
    \item[] Answer: \answerYes{} % Replace by \answerYes{}, \answerNo{}, or \answerNA{}.
    \item[] Justification: Details of the datasets, code, and findings are all available on our website. We have also provided a blog on this website with a more user-friendly explanation of the approach and findings. this aims to increase accessibility of the results to a broader audience.
    \item[] Guidelines:
    \begin{itemize}
        \item The answer NA means that the paper does not release new assets.
        \item Researchers should communicate the details of the dataset/code/model as part of their submissions via structured templates. This includes details about training, license, limitations, etc. 
        \item The paper should discuss whether and how consent was obtained from people whose asset is used.
        \item At submission time, remember to anonymize your assets (if applicable). You can either create an anonymized URL or include an anonymized zip file.
    \end{itemize}

\item {\bf Crowdsourcing and Research with Human Subjects}
    \item[] Question: For crowdsourcing experiments and research with human subjects, does the paper include the full text of instructions given to participants and screenshots, if applicable, as well as details about compensation (if any)? 
    \item[] Answer: \answerNA{} % Replace by \answerYes{}, \answerNo{}, or \answerNA{}.
    \item[] Justification: While no crowdsourcing was utilized, details of how our results are gathered and validated by the research team are provided. 
    \item[] Guidelines:
    \begin{itemize}
        \item The answer NA means that the paper does not involve crowdsourcing nor research with human subjects.
        \item Including this information in the supplemental material is fine, but if the main contribution of the paper involves human subjects, then as much detail as possible should be included in the main paper. 
        \item According to the NeurIPS Code of Ethics, workers involved in data collection, curation, or other labor should be paid at least the minimum wage in the country of the data collector. 
    \end{itemize}

\item {\bf Institutional Review Board (IRB) Approvals or Equivalent for Research with Human Subjects}
    \item[] Question: Does the paper describe potential risks incurred by study participants, whether such risks were disclosed to the subjects, and whether Institutional Review Board (IRB) approvals (or an equivalent approval/review based on the requirements of your country or institution) were obtained?
    \item[] Answer: \answerNA{} % Replace by \answerYes{}, \answerNo{}, or \answerNA{}.
    \item[] Justification: The paper does not involve crowdsourcing nor research with human subjects.
    \item[] Guidelines:
    \begin{itemize}
        \item The answer NA means that the paper does not involve crowdsourcing nor research with human subjects.
        \item Depending on the country in which research is conducted, IRB approval (or equivalent) may be required for any human subjects research. If you obtained IRB approval, you should clearly state this in the paper. 
        \item We recognize that the procedures for this may vary significantly between institutions and locations, and we expect authors to adhere to the NeurIPS Code of Ethics and the guidelines for their institution. 
        \item For initial submissions, do not include any information that would break anonymity (if applicable), such as the institution conducting the review.
    \end{itemize}
    
\item {\bf Declaration of LLM usage}
    \item[] Question: Does the paper describe the usage of LLMs if it is an important, original, or non-standard component of the core methods in this research? Note that if the LLM is used only for writing, editing, or formatting purposes and does not impact the core methodology, scientific rigorousness, or originality of the research, declaration is not required.
    %this research? 
    \item[] Answer: \answerNA{} % Replace by \answerYes{}, \answerNo{}, or \answerNA{}.
    \item[] Justification: LLM is not used as the core methodology here.
    \item[] Guidelines:
    \begin{itemize}
        \item The answer NA means that the core method development in this research does not involve LLMs as any important, original, or non-standard components.
        \item Please refer to our LLM policy (\url{https://neurips.cc/Conferences/2025/LLM}) for what should or should not be described.
    \end{itemize}
\end{enumerate}

\end{document}